\documentclass{article}
\usepackage[table]{xcolor}
\usepackage[utf8]{inputenc}
\usepackage{algorithm}
\usepackage{algpseudocode}
\usepackage{amsmath,amssymb,amsfonts}
\usepackage{arxiv}
\usepackage{cite}
\usepackage{dsfont}
\usepackage{enumitem}
\usepackage{epsfig}
\usepackage{flushend}
\usepackage{graphics}
\usepackage{graphicx}
\usepackage{hyperref}
\usepackage{lipsum}
\usepackage{listings}
\usepackage{multirow}
\usepackage{ragged2e}
\usepackage{setspace}
\usepackage{siunitx}
\usepackage{url}
\usepackage{bm}

\newcommand\scalemath[2]{\scalebox{#1}{\mbox{\ensuremath{\displaystyle #2}}}}

\usepackage{hyperref}
\hypersetup{
pdftitle={Range-Only Localization System for Small-Scale Flapping-Wing Robots},
pdfauthor={Raul Tapia, Ivan Gutierrez Rodriguez, Javier Luna-Santamaria, Jose Ramiro Martinez-de Dios, Anibal Ollero},
pdfsubject={40th Anniversary of the IEEE International Conference on Robotics and Automation}
}

\title{Range-Only Localization System for Small-Scale Flapping-Wing Robots}

\author{
\href{https://orcid.org/0000-0002-4435-5466}{\includegraphics[scale=0.06]{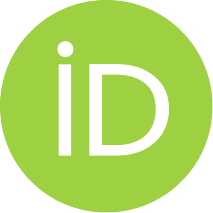}\hspace{1mm}Raul Tapia}\\
	GRVC Robotics Lab.\\
	Universidad de Sevilla\\
	\hphantom{00}\texttt{raultapia@us.es}\hphantom{00}\\
	\And
\href{https://orcid.org/0009-0009-7222-5835}{\includegraphics[scale=0.06]{orcid.pdf}\hspace{1mm}Iván Gutiérrez Rodríguez}\\
	GRVC Robotics Lab.\\
	Universidad de Sevilla\\
	\hphantom{00}\texttt{igrodriguez@us.es}\hphantom{00}\\
	\And
\href{https://orcid.org/0009-0009-5739-1894}{\includegraphics[scale=0.06]{orcid.pdf}\hspace{1mm}Javier Luna-Santamaria}\\
	GRVC Robotics Lab.\\
	Universidad de Sevilla\\
	\hphantom{00}\texttt{javierluna@us.es}\hphantom{00}\\
	\And
\href{https://orcid.org/0000-0001-9431-7831}{\includegraphics[scale=0.06]{orcid.pdf}\hspace{1mm}José Ramiro Martínez-de Dios} \\
	GRVC Robotics Lab.\\
	Universidad de Sevilla\\
	\texttt{jdedios@us.es}\\
	\And
\href{https://orcid.org/0000-0003-2155-2472}{\includegraphics[scale=0.06]{orcid.pdf}\hspace{1mm}Anibal Ollero} \\
	GRVC Robotics Lab.\\
	Universidad de Sevilla\\
	\texttt{aollero@us.es}\\
}

\paperurl{https://raultapia.com/preprints/0006/redirect}
\journal{40th Anniversary of the IEEE International Conference on Robotics and Automation}
\editorial{IEEE}

\begin{document}

\maketitle

\begin{abstract}
The design of localization systems for small-scale flapping-wing aerial robots faces relevant challenges caused by the limited payload and onboard computational resources. This paper presents an ultra-wideband localization system particularly designed for small-scale flapping-wing robots. The solution relies on custom 5 grams ultra-wideband sensors and provides robust, very efficient (in terms of both computation and energy consumption), and accurate (mean error of 0.28 meters) 3D position estimation. We validate our system using a Flapper Nimble+ flapping-wing robot.
\end{abstract}

\section{Introduction}
Accurate and robust localization plays a key role for autonomous aerial robots. Today, LiDAR-based and camera-based (indoors and outdoors) and GNSS-based (outdoors) solutions are widely used. However, the emergence of flapping-wing robots \cite{rodriguez2021griffin,tapia2023comparison,croon2020flapping} has motivated a paradigm change. First, the limited payload and the resource-constrained computation impose a limitation on the number and type of sensors to be mounted \cite{tapia2023experimental}. Second, ornithopters' flapping strokes entail several challenges for perception (e.g., motion blur in cameras) \cite{croon2010appearance}. Those restrictions are even more critical in the case of flapping-wing micro air vehicles (FWMAV) \cite{croon2009design}. Authors in \cite{ma2013controlled} use an external motion capture system to enable 6 DoF pose control. The works \cite{wagter2014autonomous} and \cite{tijmons2017obstacle} introduce an onboard stereo vision system for obstacle avoidance using FWMAVs. Range-based localization, such as those based on ultra-wideband (UWB) sensors, becomes a suitable solution for small-scale flapping-wing robots. Work \cite{grossiwindhager2019snaploc} presents an ultra-fast UWB localization system for a large number of tags. An UWB-based quadcopter localization system using range measurements is proposed in \cite{guo2016ultra}.

This paper presents a range-only localization system specifically designed for FWMAVs. It estimates the robot localization by using the distance measurements between a set of static beacons (UWB anchors at known position) and the UWB tag carried by the flapping-wing robot. Localization is computed online in a Ground Control Station (GCS) by fusing the estimates with an Extended Kalman Filter (EKF). The contributions of the paper are:

\begin{itemize}[leftmargin=*]
    \item \textbf{Hardware}. Due to the strict payload restrictions of small-scale flapping-wing robots, all components must be ultra-low weight. The localization system uses as tags and anchors the same \SI{5}{\gram} devices.
    \item \textbf{Software.} Robot-GCS communication must be robust and highly efficient due to the low computational resources. We propose a low-level communication protocol designed to work even when several anchors are lost.
\end{itemize}

We release our PCB design, UWB sensor software, and localization module implementation\footnote{\url{https://github.com/raultapia/uwb-localization}}.

\begin{figure}[ht]
\centering
\includegraphics[width=0.85\linewidth]{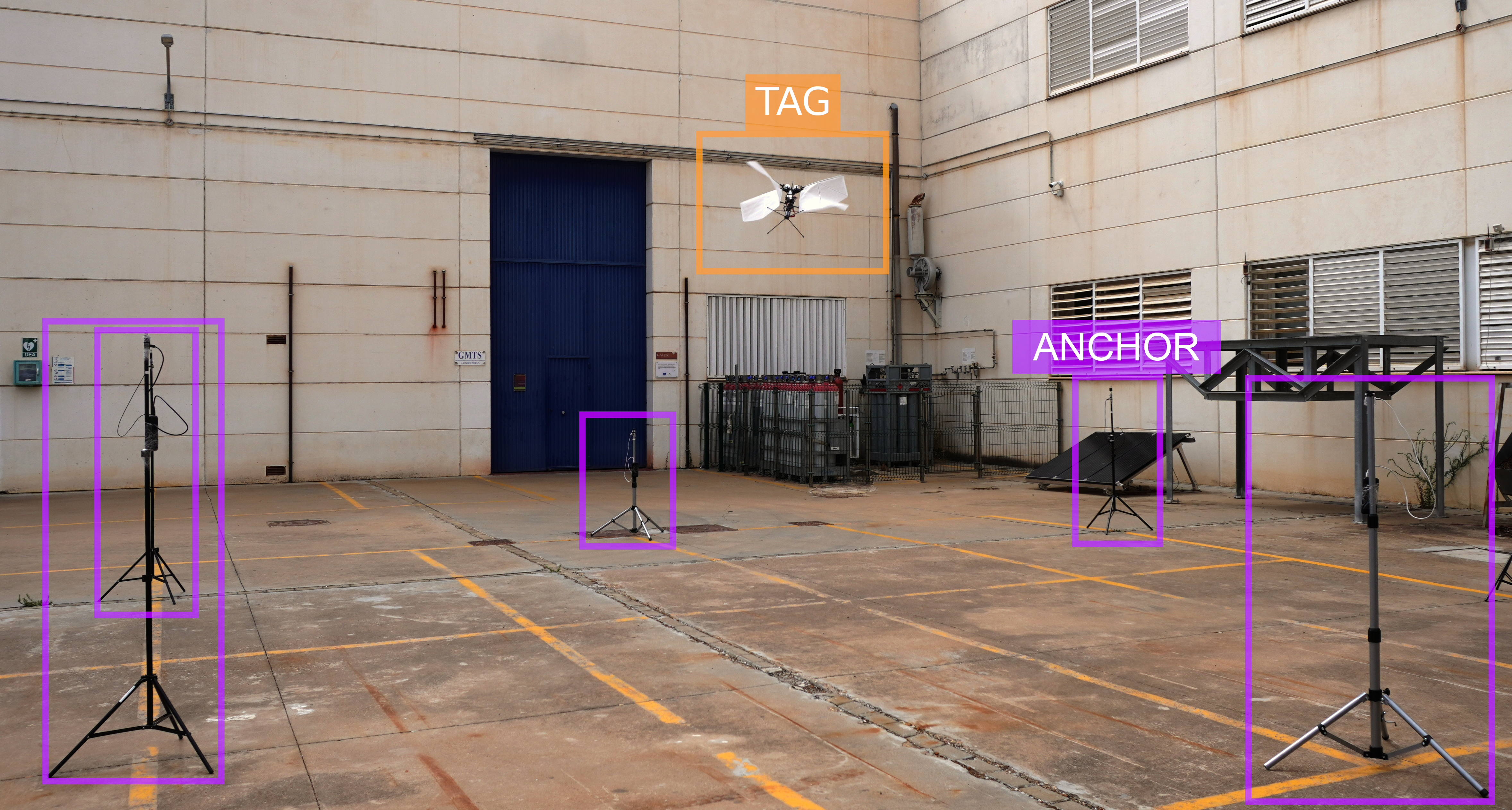}
\caption{Experimental setup used for validation. The flapping-wing robot (tag) is equipped with a UWB sensor. A total of 6 anchors are used.}
\label{fig:intro}
\end{figure}

\begin{figure}[ht]
\centering
\includegraphics[width=0.12\linewidth]{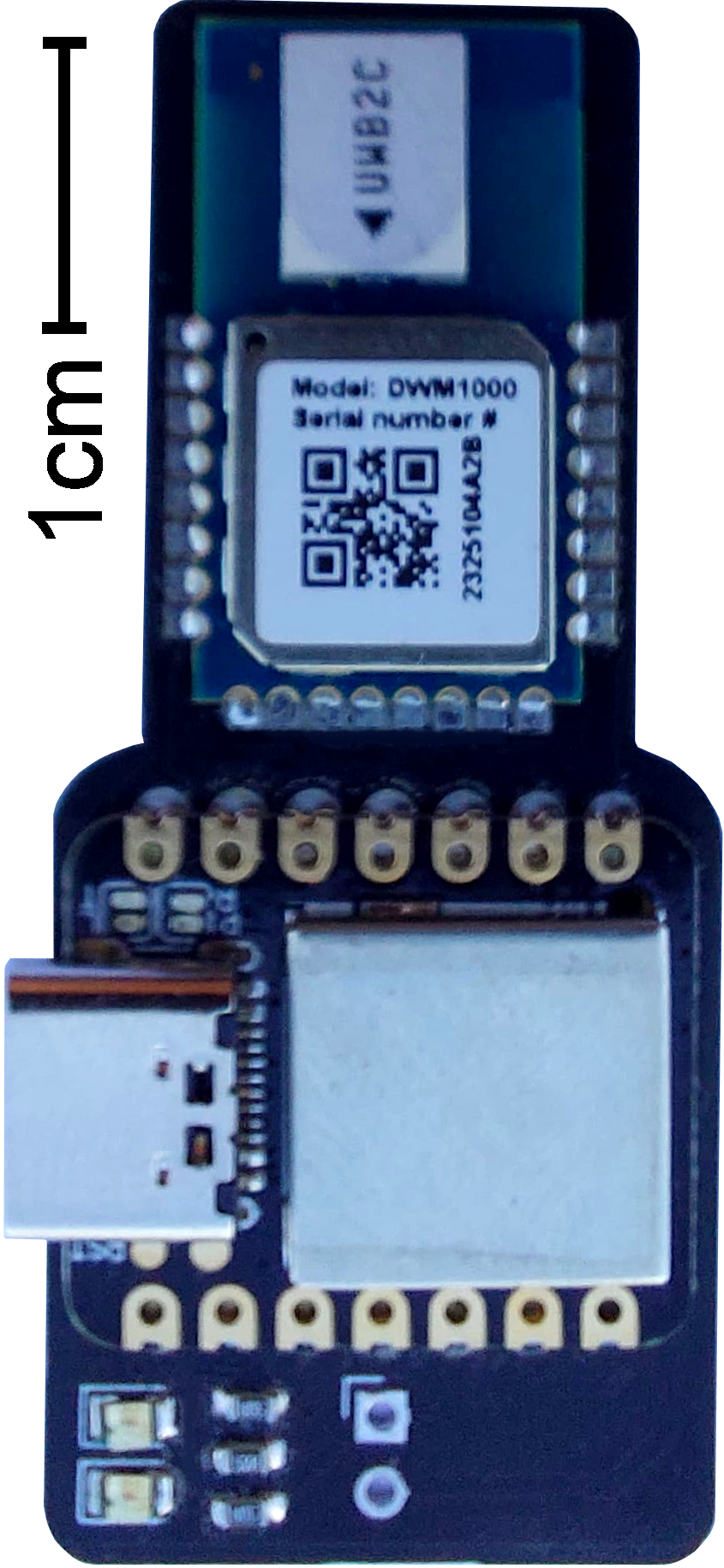}
\hspace{2cm}
\includegraphics[width=0.49\linewidth]{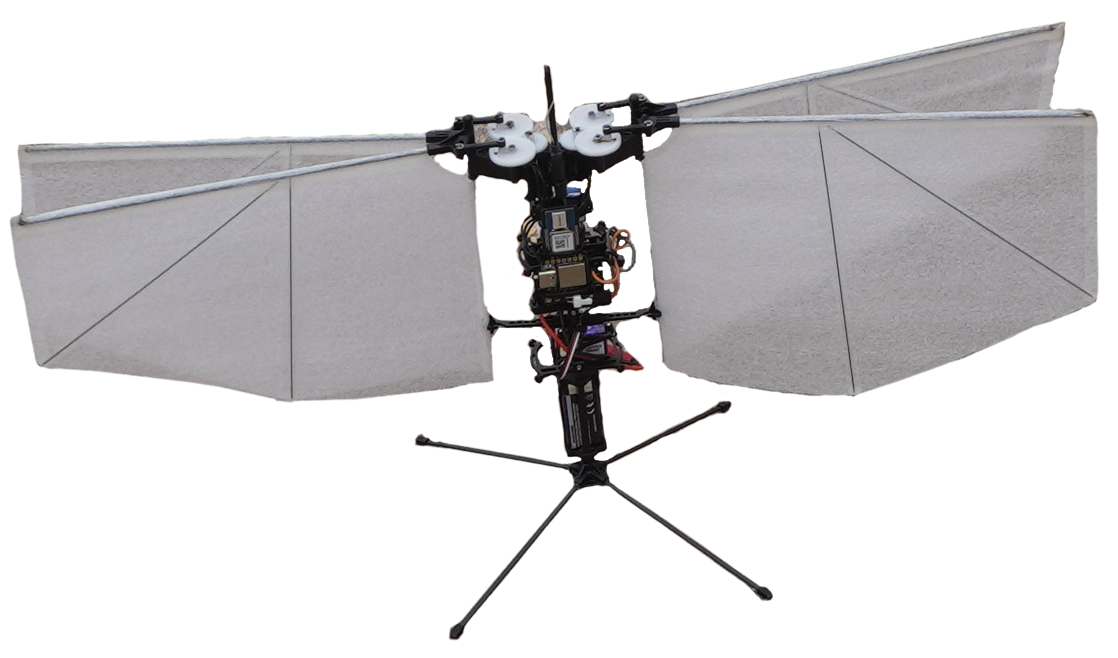}
\caption{Custom DWM1000-based UWB sensors (\SI{5}{\gram}, left) and Flapper Nimble+ (\SI{114}{\gram}, \SI{49}{\centi\meter} wingspan, right).}
\label{fig:bird}
\end{figure}

\section{Sensing and Localization}
\subsection{UWB-based Range Sensors}

\begin{figure}[ht]
\centering
\includegraphics[width=0.85\linewidth,page=1]{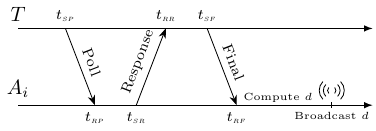}
\caption{Communication protocol between tag $T$ and anchor $A_i$ used for distance estimation.}
\label{fig:communications}
\end{figure}

The localization system relies on UWB technology due to its high accuracy and robustness against outliers. Our sensors are composed of an IEEE 802.15.4a DWM1000 UWB transceiver module operated with a Seeeduino XIAO (SAMD21G18 microcontroller). All components are mounted on a custom PCB, see \autoref{fig:bird}-left, with a total weight of \SI{5}{\gram}. The tag and the anchors use the same hardware (but different software). Our implementation is built on top of the \texttt{arduino-dw1000}\footnote{\url{https://github.com/thotro/arduino-dw1000}} library. The existing protocols for range estimation with DWM1000 include in the same message the distance measurements gathered from all the anchors. These protocols work well when using settings with 3 or 4 anchors, but require sending large messages when using a high number of anchors, compromising the communications. Based on the \textit{Two Way Ranging} method, we implemented a communication protocol to overcome this problem. The tag starts the communication with each anchor (one after the other). If an anchor does not respond to a poll after a certain time interval, it is skipped. We follow the approach illustrated in \autoref{fig:communications}. After computing the distance from the measured times\footnote{Subindex notation: $S\_$ and $R\_$ stands for \textit{send} and \textit{receive}. $\_P$, $\_R$, $\_F$ stands for \textit{poll}, \textit{response}, and \textit{final}.} ($t_{SP}$, $t_{RP}$, $t_{SR}$, $t_{RR}$, $t_{SF}$, and $t_{RF}$), the anchor broadcasts the result to the other devices. Hence, GCS can be connected to any anchor.

\subsection{Outlier Rejection}
To reject outliers, we set UWB sensors to only report range distances between \SI{0.1}{\meter} and \SI{100}{\meter}. In addition, only coherent measurements are used. A measurement is considered coherent if its error with respect to the corresponding tag-anchor distance (assuming {that the robot 3D position is accurately estimated) is less than \SI{2}{\meter}. The latter mechanism is only enabled after the convergence of the EKF.

\subsection{Localization Algorithm}
The range measurements are fused using an EKF running in GCS to estimate the robot localization. We define the system state as $\bm{\chi}= [\mathbf{x}^T, \mathbf{v}^T]^T$, where $\mathbf{x}$ and $\mathbf{v}$ are the 3D position and linear velocity of the robot UWB tag, respectively\footnote{Our method estimates position and linear velocity. In this work, we do not estimate orientation. The integration of additional sensors such as IMU for full pose estimation is considered as part of the work-in-progress.}. The EKF prediction stage is performed by assuming a constant linear velocity model (Eq. \ref{eq:kalman}-left). The EKF update stage is performed using $\mathbf{z}_k = [d_k^{(i)}, d_{k-1}^{(i)}]^T$ and the measurement model in Eq. \ref{eq:kalman}-right.
\begin{equation} \label{eq:kalman}
\scalemath{0.9}{
    \hat{\bm{\chi}}_k =
    \begin{bmatrix}
        \mathbf{x}_{k-1} + \mathbf{v}_{k-1} \Delta t_k\\
        \mathbf{v}_{k-1}
    \end{bmatrix}
    \quad
    h(\hat{\bm{\chi}}_k) =
    \begin{bmatrix}
        \|\hat{\mathbf{x}}_k-\mathbf{p}^{(i)}\|\\
        \|\hat{\mathbf{x}}_{k-1}-\mathbf{p}^{(i)}\|
    \end{bmatrix}
}
\end{equation}

\noindent
\textbf{Notation:} $\hat{a}$ stands for prediction of $a$, $\Delta t_k = t_k - t_{k-1}$, $d_k^{(i)}$ is the $k$-th distance from the $i$-th anchor, and $\mathbf{p}^{(i)}$ is the position of the $i$-th anchor (known).
\color{black}

\section{Experiments}
The method was widely experimented in settings using different number of anchors. \autoref{fig:intro} shows one picture in an experiment with 6 anchors in a \SI{9}{\meter}$\times$\SI{9}{\meter} scenario. First, as preliminary validation, the robot tag was manually moved following two predefined trajectories: \SI{4.5}{\meter}-side square (\autoref{fig:experiment}-top-left) and a trajectory describing the word \texttt{GRVC} (\autoref{fig:experiment}-top-center). Next, we used remotely controlled Flapper Nimble+\footnote{Flapper Nimble+ by Flapper Drones. [Online]. Available: \url{https://flapper-drones.com/wp/nimbleplus/}.} (\SI{114}{\gram}, \SI{49}{\centi\meter} wingspan) equipped with our UWB tag. \autoref{fig:experiment}-top-right shows the estimated trajectory. For validation, we computed the range error as the absolute difference between the distances from an anchor to the estimated robot position and the measured distance. \autoref{fig:experiment}-bottom shows the range error for each anchor in the Flapper trajectory. The mean error is $\SI{0.28}{\meter}$ with standard deviation $\SI{0.21}{\meter}$. All errors reported in all performed experiments are $<$\SI{1}{\meter}.

\begin{figure}[ht]
\centering
\includegraphics[width=0.85\linewidth, page=2]{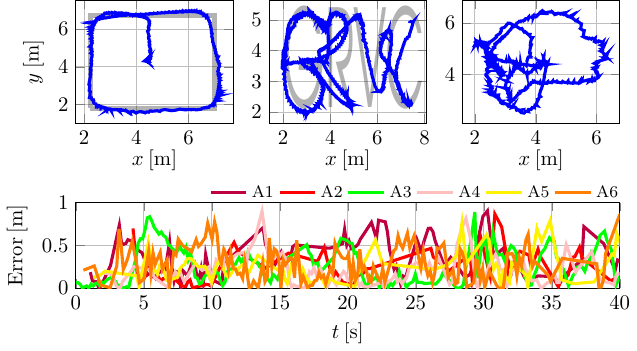}
\vspace{-1em}
\caption{Estimated trajectories in the preliminary tests (top-left, top-center) and in one real flight (top-right). The errors between the distances from the anchors to the estimated robot position and the measured distances for the real flight are shown (bottom). $Ai$ denotes the $i$-th anchor.}
\label{fig:experiment}
\end{figure}

\section{Conclusions and Work in Progress}
This paper has presented the design and implementation of a localization system for small-scale flapping-wing aerial robots based on UWB sensors. Motivated by the payload and computational constraints of FWMAVs, our solution relies on lightweight components and highly efficient communications. We are currently working on the integration of additional sensors (e.g., IMU, camera) and the extension of our framework to perform trajectory control.

\section*{Acknowledgements}
This work was funded by projects AENA-BIRD (\textit{Pájaro robótico autónomo para el ámbito aeroportuario}) funded by AENA and SARA \textit{(Sistema aéreo no tripulado seguro para la inspección de líneas eléctricas fuera de la línea de vista)} funded by Spanish Ministerio de Ciencia e Innovación. Partial funding was obtained from the Plan Estatal de Investigación Científica y Técnica y de Innovación of the Ministerio de Universidades del Gobierno de España (FPU19/04692).

\bibliographystyle{IEEEtran}
\bibliography{icra40}

\end{document}